\documentclass[11pt,a4paper]{article}
\usepackage[hyperref, acceptedWithA]{tacl2018v2}
\usepackage{times,latexsym}
\usepackage{url}
\usepackage[T1]{fontenc}
\usepackage[textsize=small]{todonotes}
\usepackage{amsmath}
\usepackage{booktabs}
\usepackage{bbm}

\DeclareMathOperator*{\argmax}{argmax}

\newcommand{\smalltablenr}[1]{\raisebox{.1\height}{\scriptsize #1}}

\title{Learning Neural Sequence-to-Sequence Models from Weak Feedback with Bipolar Ramp Loss}

\author{
	Laura Jehl\Thanks{Both authors contributed equally to this publication.} \qquad Carolin Lawrence\footnotemark[1] \\
	Computational Linguistics\\
	Heidelberg University\\
	69120 Heidelberg, Germany\\
	{\sf \{jehl, lawrence\}@cl.uni-heidelberg.de}
	\And
	Stefan Riezler\\
	Computational Linguistics \& IWR\\
	Heidelberg University\\
	69120 Heidelberg, Germany\\
	{\sf riezler@cl.uni-heidelberg.de}
	}

\date{}

\begin{document}
\maketitle

\begin{abstract}
In many machine learning scenarios, supervision by gold labels is not available and consequently neural models cannot be trained directly by maximum likelihood estimation (MLE). In a weak supervision scenario, metric-augmented objectives can be employed to assign feedback to model outputs, which can be used to extract a supervision signal for training. We present several objectives for two separate weakly supervised tasks, machine translation and semantic parsing. We show that objectives should actively discourage negative outputs in addition to promoting a surrogate gold structure. This notion of bipolarity is naturally present in ramp loss objectives, which we adapt to neural models. We show that bipolar ramp loss objectives outperform other non-bipolar ramp loss objectives and minimum risk training (MRT) on both weakly supervised tasks, as well as on a supervised machine translation task. Additionally, we introduce a novel token-level ramp loss objective, which is able to outperform even the best sequence-level ramp loss on both weakly supervised tasks.
\end{abstract}

\section{Introduction}
Sequence-to-sequence neural models are standardly trained using a maximum likelihood estimation (MLE) objective. However, MLE training requires full supervision by gold target structures,  which in many scenarios are too difficult or expensive to obtain. For example, in semantic parsing for question-answering it is often easier to collect gold answers rather than gold parses \citep[inter alia]{ClarkeETAL:10,BerantETAL:13,PasupatLiang:15,RajpurkarETAL:16}. 
In machine translation, there are many domains for which no gold references exist, however cross-lingual document-level links are present for many multilingual data collections. 

In this paper we investigate methods where a supervision signal for output structures can be extracted from weak feedback. 
In the following, we use \textit{learning from weak feedback}, or \textit{weakly supervised learning}, to refer to a scenario where output structures generated by the model are judged according to an external metric, and this feedback is used to extract a supervision signal that guides the learning process. Metric-augmented sequence-level objectives from reinforcement learning \citep{Williams:92,ranzato2016mixer}, minimum risk training (MRT) \citep{smith2006minimum, shen2016minimum} or margin-based structured prediction objectives \citep{taskar2005learning, edunov2018classical} can be seen as instances of such algorithms. 

In natural language processing applications, such algorithms have mostly been used in combination with \textit{full supervision tasks}, allowing to compute a feedback score from metrics such as BLEU or F-score that measure the similarity of output structures against gold structures. Our main interest is in \textit{weak supervision tasks} where the calculation of a feedback score cannot fall back onto gold structures. 
For example, matching proposed answers to a gold answer can guide a semantic parser towards correct parses, and matching proposed translations against linked documents can guide learning in machine translation.

In such scenarios the judgments by the external metric may be unreliable and thus unable to select a good update direction.   
It is our intuition that a more reliable signal can be produced by not just encouraging outputs that are good according to weak positive feedback, but also by actively discouraging bad structures. In this way, a system can more effectively learn what distinguishes good outputs from bad ones. We call an objective that incorporates this idea a \textit{bipolar} objective.
The bipolar idea is naturally captured by the structured ramp loss objective \cite{chapelle2009tighter}, especially in the formulation by \citet{gimpel2012ramp} and \citet{chiang2012hope}, who use 
ramp loss to separate a \textit{hope} from a \textit{fear} output in a linear structured prediction model. We employ several ramp loss objectives for two weak supervision tasks, and adapt them to neural models.

First, we turn to the task of semantic parsing in a setup where only question-answer pairs, but no gold semantic parses are given. We assume a baseline system has been trained using a small supervised data set of question-parse pairs under the MLE objective. The goal is to improve this system by leveraging a larger data set of question-answer pairs. During learning, the semantic parser suggests parses for which corresponding answers are retrieved. These answers are then compared to the gold answer and the resulting weak supervision signal guides the semantic parser towards finding correct parses. 
We can show that a bipolar ramp loss objective can improve upon the baseline by over 12 percentage points in F1 score.

Second, we employ ramp losses on a machine translation task where only weak supervision in the form of cross-lingual document-level links is available. 
We assume a translation system has been trained using MLE on out-of-domain data. 
We then investigate whether document-level links can be used as a weak supervision signal to adapt the translation system to the target domain. 
We formulate ramp loss objectives which incorporate bipolar supervision from relevant and irrelevant documents. We also present a metric which allows us to include bipolar supervision in an MRT objective. Experiments show that bipolar supervision is crucial for obtaining gains over the baseline. 
Even with this very weak supervision, we are able to achieve an improvement of over 0.4\% BLEU over the baseline using a bipolar ramp loss. 

Finally, we turn to a fully supervised machine translation task. In supervised learning, MLE training in a fully supervised scenario has also been associated with two issues. First, it can cause \textit{exposure bias} \citep{ranzato2016mixer} because during training the model receives its context from the gold structures of the training data, but at test time the context is drawn from the model distribution instead. Second, the MLE objective is agnostic to the final evaluation metric, causing a \textit{loss-evaluation mismatch} \citep{WisemanRush:2016}. 
Our experiments use a similar setup as \citet{edunov2018classical}, who apply structured prediction losses to two fully supervised sequence-to-sequence tasks, but do not consider structured ramp loss objectives. 
Like our predecessors, we want to understand if training a pre-trained machine translation model further with a metric-informed sequence-level objective will improve translation performance by alleviating the above-mentioned issues. By gauging the potential of applying bipolar ramp loss  in a full supervision scenario, we achieve best results for a bipolar ramp loss, improving the baseline by over 0.4\% BLEU.

In sum, we show that bipolar ramp loss is superior to other sequence-level objectives for all investigated tasks, supporting our intuition that a bipolar approach is crucial where strong positive supervision is not available.  
In addition to adapting the ramp loss objective to weak supervision, our ramp loss objective can also be adapted to operate at the token level, which makes it particularly suitable for neural models as they produce their outputs token by token. A token-level objective also better emulates the behavior of the ramp loss for linear models, which only update the weights of features that differ between hope and fear. Finally, the token-level objective allows us to capture token-level errors in a setup where MLE training is not available. 
Using this objective, we obtain additional gains on top of the sequence-level ramp loss for weakly supervised tasks. 

\section{Related Work} 
Training neural models with metric-augmented objectives has been explored for various NLP tasks in supervised and weakly supervised scenarios.
MRT for neural models has previously been employed for machine translation \citep{shen2016minimum} and semantic parsing \citep{LiangETAL:17,GuuETAL:17}.\footnote{Note that \citet{LiangETAL:17} refer to their objective as an instantiation of REINFORCE, however they build an average over several outputs for one input and thus the objective more accurately falls under the heading of MRT.} Other objectives based on classical structured prediction losses have been used for both machine translation and summarization \citep{edunov2018classical}, as well as semantic parsing \citep{IyyerETAL:17,MisraETAL:18}. Objectives inspired by REINFORCE have, for example, been applied to machine translation \citep{ranzato2016mixer, nourouzi2016raml}, semantic parsing \citep{LiangETAL:17,MouETAL:17a,GuuETAL:17} and reading comprehension \citep{ChoiETAL:17, YangETAL:17}.\footnote{We do not use REINFORCE because its updates are based on only one sampled model output, which can lead to high variance. Since it is possible for us to obtain feedback for more than one model output, we employ the more robust MRT that calculates an average over several outputs.}

\citet{MisraETAL:18} are the first to compare several objectives for neural semantic parsing. For semantic parsing, they find that objectives employing structured prediction losses perform best. 
\citet{edunov2018classical} compare different classical structured prediction objectives including MRT on a fully supervised machine translation task. They find MRT to perform best. However, they only obtain larger gains by interpolating MRT with the MLE loss.
Neither \citet{MisraETAL:18} nor \citet{edunov2018classical} investigate objectives that correspond to the bipolar ramp loss that is central in our work.

The ramp loss objective \cite{chapelle2009tighter} has been applied to supervised phrase-based machine translation \citep{gimpel2012ramp,chiang2012hope}. We adapt these objectives to neural models and adapt them to incorporate bipolar weak supervision, while also introducing a novel token-level ramp loss objective.

\section{Neural Sequence-to-Sequence Learning} \label{sec:seq}
Our neural sequence-to-sequence models employ an encoder-decoder setup \citep{ChoETAL:14,SutskeverETAL:14} with an attention mechanism \citep{BahdanauETAL:15}. Specifically, we employ the framework \textsc{Nematus} \citep{SennrichETAL:17}. 
Given an input sequence $x = x_1, x_2, \dots x_{|x|}$, the probability that a model assigns for an output sequence $y = y_1, y_2, \dots y_{|y|}$ is given by
$
\pi_w(y|x) = \prod_{j=1}^{|y|} \pi_w(y_j|y_{<j},x).
$
Using beam search, we can obtain a sorted $k$-best list $\mathcal{K}(x)$ of most likely to least likely outputs and we define the most likely output as 
$
  \hat{y} = \argmax_{y \in \mathcal{K}(x)} \pi_w(y|x).
$

\paragraph{Maximum Likelihood Estimation (MLE).} Prior to employing metric-augmented objectives, we assume that a model has been pre-trained with a maximum likelihood estimation (MLE) objective. Given inputs $x$ and gold structures $\bar{y}$, the parameters of the neural network are updated using Stochastic Gradient Descent (SGD) with minibatches of size $M$, leading to the following objective:
\begin{align}
\mathcal{L}_{MLE} =
- \frac{1}{M} \sum_{m=1}^{M} \sum_{j=1}^{|\bar{y}|} \log \pi_w(\bar{y}_{m,j} | \bar{y}_{m,<j}, x_m).
\end{align}

\paragraph{Minimum Risk Training (MRT).} We compare our ramp loss objectives to MRT \citep{shen2016minimum}, which employs an external metric to assign rewards to model outputs. Given an input $x$, $S$ outputs are sampled from the model distribution and updates are performed based on the following MRT objective:
\begin{equation}
\mathcal{L}_{\textsc{MRT}} = - \frac{1}{M} \sum_{m=1}^M \frac{1}{S} \sum_{s=1}^{S} \pi_w(y_{m,s}|x_m) \delta(y_{m,s}),
\end{equation}
where $\delta(y_{m,s})$ is the reward returned for $y_{m,s}$ by the external metric, and $\pi_w(y_{m,s}|x_m)$ is a distribution over outputs that is normalized over $S$ samples and can be controlled for sharpness by a temperature parameter.\footnote{We follow the implementation of MRT in \textsc{Nematus} with its default settings, including de-duplication of samples and setting the temperature parameter to $\alpha=0.005$. In case of fully supervised MT where the question arises whether to include the reference in the sample, we choose not to include it in order to be comparable with \citet{edunov2018classical} who also do not include it.}
Following  \citet{shen2016minimum}, we use a baseline term $b(x_m)$ that acts as a control variate for variance reduction of the stochastic gradient \cite{Williams:92,GreensmithETAL:04} and allows negative updates for rewards smaller than the baseline. We compute this term by sampling $S^\prime$ outputs from the model distribution s.t.
$
b(x) = -\frac{1}{S^\prime} \sum_{s^\prime=1}^{S^\prime} \delta(y_{s^\prime}).
$

 \paragraph{Ramp Loss Objectives.}

Our ramp loss objectives can be formulated as follows:
\begin{align}
\label{eq:ramp_loss}
\mathcal{L}_{\textsc{RAMP}} =& \phantom{-}  \frac{1}{M} \sum_{m=1}^{M} \pi_w(y_m^-|x_m)\\ \notag
&-  \frac{1}{M} \sum_{m=1}^{M} \pi_w(y_m^+|x_m),
\end{align}
where $y^-$ is a \textit{fear} output that is to be discouraged and $y^+$ is a \textit{hope} output that is to be encouraged. Intuitively, $y^-$ should be an output which has high probability, but receives a bad reward from the external metric. Analogously, $y^+$ should be an output which has high probability and receives a high reward from the external metric. The concrete instantiations of $y^-$ and $y^+$ depend on the underlying task and are thus deferred to the respective sections below (see Tables \ref{tab:ramp_parse}, \ref{tab:ramp_weakmt} and \ref{tab:ramp_fullmt}). The \textsc{RAMP} loss defined in equation \eqref{eq:ramp_loss} has been introduced as equation (8) in \citet{gimpel2012ramp}. This loss naturally incorporates a bipolarity principle by including both \textit{hope} and \textit{fear} into one objective. An alternative formulation of ramp loss can be given by favoring the current model prediction, i.e., setting $y^+=\hat{y}$, and searching for a \textit{fear} output. This has been called ``cost-augmented decoding'' and been formalized in equation (6) in \citet{gimpel2012ramp}. This loss  dates back to the ``margin-rescaled hinge loss'' of \citet{TaskarETAL:04} and will be called \textsc{RAMP1} in the following. The converse approach has been called ``cost-diminished decoding'' and been formalized in equation (7) in \citet{gimpel2012ramp}. Here the model prediction is penalized by setting $y^-=\hat{y}$ and searching for a \textit{hope} output. This objective has been called ``direct loss'' in \citet{mcallester2010direct}, and will be called \textsc{RAMP2} in the following.

Finally, we introduce a ramp loss objective which can operate on the token level. To be able to adjust individual tokens, we move to $\log$ probabilities, so that the sequence decomposes as a sum over individual tokens and it is possible to ignore tokens while encouraging or discouraging others. 
This leads to the \textsc{Ramp-T} objective:
\begin{align}
\label{eq:loss_token}
&\mathcal{L}_{\textsc{Ramp-T}} =\\ \notag
& \phantom{-} \frac{1}{M} \sum_{m=1}^{M} \sum_{j=1}^{|y_m^-|} \tau^-_{m,j} \log \pi_w(y_{m,j}^-|y_{m,<j},x_m)\\ \notag
& - \frac{1}{M} \sum_{m=1}^{M} \sum_{j=1}^{|y_m^+|} \tau^+_{m,j} \log \pi_w(y_{m,j}^+|y_{m,<j},x_m),
\end{align}
where $\tau^+_{m,j}$ and $\tau^-_{m,j}$ are set to $0$, $1$ or $-1$ depending on the decision whether the corresponding token $y^+_{m,j} / y^-_{m,j}$ should be left untouched, encouraged or discouraged. Concretely, we define:
\begin{equation}
\tau^+_{m,j} = \begin{cases}
0&\text{if}\, y^+_{m,j} \in y^-\\
1&\text{else}
\end{cases}	\end{equation}
and 
\begin{equation}\tau^-_{m,j} = \begin{cases}
\phantom{-} 0&\text{if}\, y^-_{m,j} \in y^+\\
-1&\text{else.}
\end{cases}	\end{equation}

\begin{figure}[t]
	\centerline{\includegraphics[width=0.25\textwidth,keepaspectratio]{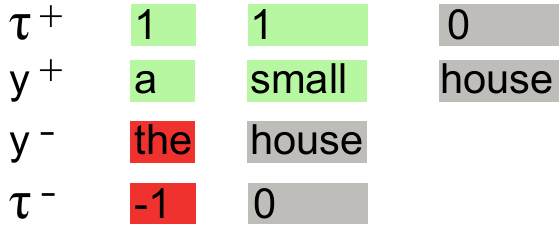}}
	\caption{Settings for token-level rewards $\tau+$ and $\tau^-$ for hope output $y^+$ = ``\textit{a small house}'' and fear output $y^-$ = ``\textit{the house}''.}
	\label{fig:token_level}
\end{figure}

With this definition, tokens that appear in both $y^+$ and $y^-$ are left untouched, whereas tokens that appear only in the hope output are encouraged, and tokens that appear only in the fear output are discouraged (see Figure \ref{fig:token_level} for an example). This more fine-grained contrast allows the model to learn what distinguishes a good output from a bad one more effectively.\footnote{An implementation of the \textsc{Ramp} objectives can be found at \url{https://github.com/carhaas/nematus}.}

\section{Semantic Parsing}

\paragraph{Ramp Loss Objectives.}
In semantic parsing for question answering, natural language questions are mapped to machine readable parses. Such a parse, $y$, can be executed against a database which returns an answer $a$. This answer $a$ can be compared to the available gold answer $\bar{a}$ and the following metric can be defined:
\begin{equation}
\label{eq:feedback}
\delta(y) = \begin{cases}
1&\text{if}\enspace a = \bar{a}\\
0&\text{else.}
\end{cases}
\end{equation}

\begin{table*}
	\begin{center}
		\begin{tabular}{llll}
			\toprule
			Name& $y ^+$ & $y^-$\\
			\midrule
			\textsc{RAMP}&$\argmax_{y \in \mathcal{P}(x)} \pi_w(y|x)$&$\argmax_{y \in \mathcal{N}(x)} \pi_w(y|x)$\\
			\textsc{RAMP1}&$\hat{y}$&$\argmax_{y \in \mathcal{N}(x)} \pi_w(y|x)$\\
			\textsc{RAMP2}&$\argmax_{y \in \mathcal{P}(x)} \pi_w(y|x)$&$\hat{y}$\\
			\bottomrule
		\end{tabular}
		\caption{Configurations for $y^+$ and $y^-$ for semantic parsing. We abbreviate $\mathcal{P}(x)=\mathcal{K}(x):\delta(y)=1$, which is the most likely output in the $k$-best list $\mathcal{K}(x)$ that leads to the correct answer, and $\mathcal{N}(x)=\mathcal{K}(x):\delta(y)=0$, which is the most likely output in the $k$-best list $\mathcal{K}(x)$ that leads to the wrong answer.}
		\label{tab:ramp_parse}
	\end{center}
\end{table*}

For \textsc{RAMP}, $y^+$ is defined as the most probable output in the $k$-best list $\mathcal{K}(x)$ that leads to the correct answer, i.e.~where $\delta(y)=1$. In contrast, $y^-$ is defined as the most probable output in $\mathcal{K}(x)$ that does not lead to the correct answer, i.e.~where $\delta(y)=0$. The definitions of $y^+$ and $y^-$ for this objective and the related ramp loss objectives can be found in Table \ref{tab:ramp_parse}. If $y^+$ or $y^-$ are found, the parse is cached as a hope or fear output, respectively, for the corresponding input $x$. If at a later point $y^+$ or $y^-$ cannot be found in the current $k$-best list, then previously cached outputs are accessed instead. Should no cached output exist, the corresponding sample is skipped.

\paragraph{Experimental Setup.}
Our experiments are conducted on the \textsc{NLmaps v2} corpus \citep{LawrenceRiezler:18} which is a publicly available corpus\footnote{\url{https://www.cl.uni-heidelberg.de/statnlpgroup/nlmaps/}} for geographical questions that can be answered with the \textsc{OpenStreetMap} database.\footnote{\url{https://www.openstreetmap.org}} The corpus is a recent extension of its predecessor \cite{HaasRiezler:16} which has been used in \citet{KociskyETAL:16} or \citet{DuongETAL:18}.

For each question, the corpus provides both gold parses and gold answers that can be obtained by executing the parses against the database. We take a random subset of 2,000 question-parse pairs to train an initial model $\pi_w$ with the \textsc{MLE} objective. Following \citet{LawrenceRiezler:18}, we take a pre-order traversal of the tree-structured parses to obtain individual tokens. 1,843 and 2,000 further instances of the corpus are retained for development and test set, respectively. For the remaining 22,766 questions, we assume that no gold parses exist and only gold answers are available. With the gold answers as a guide, the initial model $\pi_w$ is further improved using the metric-augmented objectives of Section \ref{sec:seq} and the metric defined in equation (\ref{eq:feedback}).

The model has 1,024 hidden units (GRUs) and word embeddings of size 1,000. The optimal learning rate was chosen in preliminary experiments on the development set and is set to $0.1$. Gradients are clipped to 1.0 if they exceed a value of 1.0 and the sentence length is capped at 200. In the case of the \textsc{MRT} objectives, we set $S=S^\prime=10$. For the \textsc{RAMP} objectives the size of the $k$-best list $\mathcal{K}$ is 10. For objectives with minibatches, the size of a minibatch is $M=80$ and validation on the development set is performed after every 100 updates. For objectives where updates are performed after each seen input, the validation is run after every 8,000 updates, leading to the same number of seen inputs compared to the objectives with minibatches.

For validation and at test time, the most likely parse is obtained after a beam search with a beam of size 12. The obtained parse is executed against the database to retrieve its corresponding answer which is compared to the available gold answer. We define recall as the percentage of correct answers in the entire set and precision as the percentage of correct answers in the set of non-empty answers. The harmonic mean of recall and precision constitutes the F1 score. The stopping point is determined by the highest F1 score on the development set after 30 validations or 30 days or run time\footnote{The 30 day mark was only hit by \textsc{Ramp2}.} and corresponding results are reported on the test set. To measure statistical significance between models we employ an approximate randomization test \citep{Noreen:1989}.

\paragraph{Experimental Results.}
Results using the various ramp loss objectives as well as \textsc{MRT} are shown in Table \ref{exp:parse_results}. \textsc{MRT} outperforms the \textsc{MLE} baseline by about 6 percentage points in F1 score. \textsc{RAMP1} performs worse than \textsc{MRT}, but can still significantly outperform the baseline by 3.05 points in F1 score. \textsc{RAMP2} performs better than \textsc{RAMP1}, but outperforms \textsc{MRT} only nominally.

In contrast to this, by carefully selecting both a hope and fear parse, \textsc{RAMP} achieves a significant further 5.43 points in F1 score over \textsc{MRT}. By incorporating token-level feedback, our novel objective \textsc{RAMP-T} outperforms all other models significantly and beats the baseline by over 12 points in F1 score. Compared to \textsc{RAMP}, \textsc{RAMP-T} can take advantage of the token-level feedback which allows a model to determine which tokens in the hope output are instrumental to obtain a positive reward but are missing in the fear output. Analogously it is possible to identify which tokens in the fear output lead to an incorrect parse, rather than also punishing the tokens in the fear output which are actually correct.

\begin{table}
	\begin{center}
		\begin{tabular}{lllll}
			\toprule
			&&$M$& \% F1 & $\Delta$ \\
			\midrule
			\raisebox{.1\height}{\scriptsize 1}&\textsc{MLE}&&57.45&\\
			\raisebox{.1\height}{\scriptsize 2}&\textsc{MRT}&$\phantom{0}$1&63.60\small$\pm0.02$&+$\phantom{0}$6.15\\
			\raisebox{.1\height}{\scriptsize 3}&\textsc{RAMP1}&80&60.50\small$\pm0.01$&+$\phantom{0}$3.05\\
			\raisebox{.1\height}{\scriptsize 4}&\textsc{RAMP2}&80&64.22\small$\pm0.00$&+$\phantom{0}$6.77\\
			\raisebox{.1\height}{\scriptsize 5}&\textsc{RAMP}&80&69.03\small$\pm0.04$&+11.58\\
			\raisebox{.1\height}{\scriptsize 6}&\textsc{RAMP-T}&80&69.87\small$\pm0.02$&+12.42\\
			\bottomrule
		\end{tabular}
		\caption{Answer F1 scores on the \textsc{NLmaps\index{NLmaps} v2} test set for various objectives, averaged over two independent runs. $M$ is the minibatch size. All models are statistically significant from each other at $p<0.01$, except the pair (2, 4).}
		\label{exp:parse_results}
	\end{center}
\end{table}

\textsc{MRT} is not naturally a bipolar objective. It can only discourage wrong parses if the baseline is larger than 0. Investigating the value of the baseline for 10,000 instances shows that in 37\% of the cases the baseline is 0, i.e. none of the sampled parses leads to the correct answer. As a result, 37\% of the time, wrong parses are ignored rather than discouraged. To explore the importance of always discouraging wrong parses, we introduce the objective \textsc{MRT neg}: it modifies the feedback for parses with a wrong answer to be $-1$ rather than $0$, which resembles the fear output that is discouraged in the \textsc{RAMP} objective. With this change, the MRT objective always behaves in a bipolar manner, irrespective of the baseline's value.
As a consequence, \textsc{MRT neg} can significantly outperform \textsc{MRT} by 2.33 points in F1 score (see Table \ref{exp:parse_results_ablation}). 
This showcases the importance of employing bipolar supervision and it constitutes an important finding compared to previous approaches \citep{LiangETAL:17,MisraETAL:18}, where the feedback is defined to lie in the range of $[0,1]$. 

However, \textsc{MRT neg} still falls short of \textsc{RAMP} by 3.1 points in F1 score. This could be because of the different batch sizes as \textsc{MRT} uses a batch size of 1, whereas \textsc{RAMP} employs a batch size of 80. To ensure that the difference between the objectives does not stem from this difference, we run an experiment with \textsc{RAMP} where the batch size is also set to 1, i.e. \textsc{RAMP m=1}. Crucially, it still significantly outperforms \textsc{MRT}.
At the same time, it does however have a lower F1 score than \textsc{RAMP} (see Table \ref{exp:parse_results_ablation}). This showcases the importance of using a larger minibatch size, so that an average over several inputs is computed before updating. In fact, its F1 score is on par with the \textsc{MRT neg} objective, which uses the same minibatch size and incorporates bipolar supervision just as \textsc{Ramp} does. However, \textsc{RAMP m=1} should still be preferred because the \textsc{RAMP} objectives are more efficient than \textsc{MRT} objectives.
In the case of \textsc{MRT}, for every training instance $S + S^\prime = 20$ queries need to be executed against the database to obtain an answer and corresponding reward. On the other hand, \textsc{Ramp} has to execute \textit{at most} the $10$ queries of the $k$-best list $\mathcal{K}$, but often less if both a correct and an incorrect query are found earlier.

\begin{table}
	\begin{center}
		\begin{tabular}{lllll}
			\toprule
			&&$M$& \% F1 & $\Delta$ \\
			\midrule
			\raisebox{.1\height}{\scriptsize 1}&\textsc{MLE}&&57.45&\\
			\raisebox{.1\height}{\scriptsize 2}&\textsc{MRT}&$\phantom{0}$1&63.60\small$\pm0.02$&+$\phantom{0}$6.15\\
			\raisebox{.1\height}{\scriptsize 3}&\textsc{MRT neg}&$\phantom{0}$1&65.93\small$\pm0.16$&+$\phantom{0}$8.48\\
			\raisebox{.1\height}{\scriptsize 4}&\textsc{RAMP m=1}&$\phantom{0}$1&66.78\small$\pm0.21$&+$\phantom{0}$9.33\\
			\raisebox{.1\height}{\scriptsize 5}&\textsc{RAMP}&80&69.03\small$\pm0.04$&+11.58\\
			\bottomrule
		\end{tabular}
		\caption{Answer F1 scores on the \textsc{NLmaps v2} test set for \textsc{RAMP} and the \textsc{MRT} objective as well as two further objectives, which help crystallize the difference between the two former objectives, averaged over two independent runs. $M$ is the minibatch size. All models are statistically significant from each other at $p<0.01$, except the pair (3, 4).}
		\label{exp:parse_results_ablation}
	\end{center}
\end{table}
To summarize, \textsc{RAMP} can attribute its success to two factors: First, it discourages parses that receive a wrong answer rather than ignoring them as \textsc{MRT} often does. Second, a larger minibatch size leads to improvements because updates are based on an average over several inputs. Further performance gains can be obtained by employing the token-level objective \textsc{RAMP-T}. Finally, \textsc{RAMP} objectives are more efficient because fewer outputs have to be judged.

\section{Weakly Supervised Machine Translation}
\label{sec:mt_weak}

\begin{table*}
\centering
\begin{tabular}{lll}
\toprule
 Loss & $y^+$ & $y^-$\\
 \midrule
\textsc{RAMP}& $\argmax_{y} \pi_w(y|x) - \alpha(1 - \delta_1(y,d^+))$  & $\argmax_{y} \pi_w(y|x) + \alpha ( 1 - \delta_1(y,d^+))$\\
\textsc{RAMP$^-$}  & $\argmax_{y} \pi_w(y|x) - \alpha(1 - \delta_1(y,d^+))$  & $\argmax_{y} \pi_w(y|x) - \alpha ( 1 - \delta_1(y,d^-))$\\ 
\textsc{RAMP1$^-$}  & $\hat{y}$ & $\argmax_{y} \pi_w(y|x) - \alpha ( 1 - \delta_1(y,d^-))$\\
\textsc{RAMP2}  & $\argmax_{y} \pi_w(y|x) - \alpha(1 - \delta_1(y,d^+))$  & $\hat{y}$\\  
\textsc{RAMP$_{\delta_2}$} & $\argmax_{y}  \pi_w(y|x) - \alpha (1 - \delta_2(y,d^+, d^-))$  & $\argmax_{y} \pi_w(y|x) + \alpha( 1 - \delta_2(y,d^+,d^-))$\\ 
\bottomrule 
\end{tabular}
\caption{Configurations for $y^+$ and $y^-$ for weakly supervised MT adaptation. $\hat{y}$ is the highest-probability model output. $\pi_w(y|x)$ is the probability of $y$ under the model. The $\argmax_y$ is  taken over the $k$-best list $\mathcal{K}(x)$. 
  $\alpha$ is a scaling factor regulating the influence of the metric compared to the model probability. $\delta_1$ and $\delta_2$ are metrics defined with respect to relevant and irrelevant documents $d^+$ and $d^-$ (see Eq. \ref{eq:mtweak_delta1} and \ref{eq:mtweak_delta2}). }
\label{tab:ramp_weakmt}
\end{table*}

\paragraph{Ramp Loss Objectives.} We consider machine translation (MT) in a weakly supervised domain adaptation setting, where in-domain references are unavailable. 
In this setting, we obtain weak feedback by matching translation model outputs against cross-lingually linked documents. 
For each input sentence $x$, we can obtain a set of \textit{relevant} documents $D^+(x) \in D$ where $D$ is a collection of target language documents. Cross-lingual link structures can be found in many multilingual document collections, such as cross-lingual citations in patent documents or product categories in e-commerce data. Our example is links between Wikipedia documents. Instead of a reference translation, we use a relevant document $d^+$ sampled from $D^+(x)$ to guide our search for $y^+$ and $y^-$. As a relevant document provides much weaker supervision than a reference translation, we construct a more informative supervision signal by integrating negative supervision from an irrelevant document $d^-$ sampled from a collection of irrelevant contrast documents. 
For each input $x$, the bipolar supervision signal then consists of a pair of sampled documents $( d^+, d^-)$. 

Unlike semantic parsing for question answering, our task uses a continuous reward $\delta(y) \in [0,1]$. 
In fully supervised MT a sentence-level approximation of the BLEU score can serve as the reward. But computing the BLEU score between a translation and a document does not make sense. We therefore propose two different alternative metrics. The first, $\delta_1(y,d)$, computes how well a translation matches a relevant document. The second, $\delta_2(y,d^+, d^-)$ computes how well a translation differentiates between a relevant and an irrelevant document.
$\delta_1(y,d)$ is defined as the average $n$-gram precision between a hypothesis and a document, multiplied by a brevity penalty. As we do not have a reference length, we include a brevity penalty term which compares the output length to the input length. This ratio can be modified by a factor $r$ that represents the average length difference between source and target language and which can be computed over the training data: 
\begin{equation}\label{eq:mtweak_delta1}
	\delta_1(y,d) = \frac{1}{N} \sum_{n=1}^N \frac{\sum_{u_n}c(u_n,y) \cdot \mathbbm{1}_{u_n \in d}}{ \sum_{u_n} c(u_n,y)}
 \cdot  BP\,, 
\end{equation}
 where $u_n$ are the $n$-grams present in $y$, $c()$ counts the occurrences of an $n$-gram in $y$ and $N$ is the maximum order of $n$-grams used. The brevity penalty term is 
 \[
BP= \min (1,\frac{r \cdot |y|}{|x|})\,.
 \]
$\delta_2(y,d^+, d^-)$ is defined as the difference between $\delta_1(y,d^+)$ and $\delta_1(y, d^-)$, subject to a linear transformation to allow values to lie between 0 and 1: 
\begin{equation}\label{eq:mtweak_delta2}
\begin{split}
&\delta_2(y,d^+, d^-) = \\
&	0.5 \cdot (\delta_1(y,d^+) - \delta_1(y,d^-) + 1)\,.
\end{split}
\end{equation}
Our intuition behind this metric is that it should measure how well a translation differentiates between the relevant and irrelevant document, leading to domain-specific translations being weighted higher than domain-agnostic ones.

Table \ref{tab:ramp_weakmt} shows our loss functions for the weakly supervised case. 
RAMP and RAMP2 define $y^+$ and $y^-$ in the same way as is done in the semantic parsing task, except that the metric $\delta_1(y,d^+)$ is employed to match outputs against documents. Like \citet{gimpel2012ramp}, we include a scaling factor $\alpha$ to trade off the importance of the reward against the model score in determining $y^+$ and $y^-$. 
Note that these objectives do not include negative supervision from $d^-$. 
Using the metrics defined above, we formulate two objectives that include $d^-$:
RAMP$^-$ defines $y^+$ in the same way as RAMP, but uses a different definition of $y^-$: Instead of using a \textit{fear} output with respect to $d^+$, i.e.~a translation with high probability and low reward $\delta_1(y,d^+)$, we use a \textit{hope} output with respect to $d^-$, i.e.~a translation with high probability and high reward $\delta_1(y,d^-)$. As this translation matches an irrelevant document well, it can be used as a negative output. The same definition of $y^-$ is also used in \textsc{RAMP1$^-$}. Note that this objective does not include positive supervision from $d^+$. 
Finally, RAMP$_{\delta_2}$ incorporates $d^+$ and $d^-$ in a different way. This objective defines $y^+$ as a hope and $y^-$ as a fear, but uses the joined metric $\delta_2(y,d^+, d^-)$  with respect to the document pair $(d^+, d^-)$. 

\paragraph{Experimental Setup.}
We test our objectives on a weakly supervised English-German Wikipedia translation task first proposed in \citet{jehl2016coling}. 
In-domain training data are 10,000 English sentences with relevant German documents sampled from the WikiCLIR corpus \citep{schamoni2014}.\footnote{WikiCLIR annotates both a stronger \textit{mate} relation when there is a direct cross-lingual link between documents and a weaker \textit{link} relation when a there is a bidirectional link between a German mate document and another German document. The experiments reported here use the \textit{mate} relation.}
The task includes a small in-domain development and test set (dev: 1,712 sentences, test: 1,526 sentences), each consisting of four Wikipedia articles with diverse subjects.
Irrelevant documents $d^-$ are sampled from the German side of the News Commentary\footnote{\url{http://casmacat.eu/corpus/news-commentary.html}} data set, which contains document boundary information.
 
Byte-pair encoding \citep{sennrich2016neural} with 30,000 merge operations is applied to all source and target data. Sentences longer than 80 words are removed from the training set. 
Our neural MT model uses 500-dimensional word embeddings and hidden layer dimension of 1,024. Encoder and decoder use GRU units. 
An out-of-domain model is trained on 2.1 million sentence pairs from Europarl v7 \citep{koehn2005europarl}, News Commentary v10 and the MultiUN v1 corpus \citep{eisele2010multiun}.
The baseline (MLE) is trained using the MLE objective and \textsc{ADADELTA} \citep{zeiler2012adadelta} for 20 epochs. We train on batches of 64  and use dropout for regularization, with a dropout rate of  0.2 for embedding and hidden layers and 0.1 for source and target layers. 
Gradients are clipped if their norm exceeds 1.0.   

The metric-augmented objectives are trained using SGD. All hyperparameters are chosen on the development set. For the ramp loss objectives, we use a learning rate of 0.005, $\alpha=10$ and a $k$-best size of 16. We compare ramp loss to MRT using both $\delta_1(y,d^+)$ and $\delta_2(y,d^+,d^-)$ as the external cost function, denoted as MRT$_{\delta_1}$ and MRT$_{\delta_2}$ respectively.
MRT is trained using a learning rate of 0.05, $S=16$ and $S^\prime=10$. 
For testing and validation, translations are obtained using beam search with a beam size of 16. 
Results are validated every 200 updates and training is run for 25 validations. 
The stopping point is determined by the BLEU score \citep{Papineni:01} on the development set. We report scores computed with Moses'\footnote{\url{https://github.com/moses-smt/mosesdecoder}} \texttt{multi-bleu.perl} on tokenized, truecased output. Results are averaged over 2 runs.  

\begin{table}
\begin{center}
\begin{tabular}{cllll}
\toprule
& & $M$ & \% BLEU  & $\Delta$ \\
\midrule
\smalltablenr{1} & MLE & 64 &  15.59  & \\ 
\smalltablenr{2} & \textsc{RAMP} & 40 &  15.03 $\pm 0.01$  & $-$ 0.56 \\ 
\smalltablenr{3} & \textsc{RAMP1$^-$} & 40 & 15.12 $\pm 0.02$ & $-$ 0.47 \\ 
\smalltablenr{4} &  \textsc{RAMP2} &40 &  15.19 $\pm 0.01$ & $-$ 0.40 \\  
\smalltablenr{5} &  \textsc{MRT}$_{\delta_1}$ & 1 & 15.37 \small $\pm 0.04$ & $-$ 0.22\\ 
\smalltablenr{6} &  \textsc{MRT}$_{\delta_2}$ & 1 & 15.70 \small $\pm 0.04$ & $+$ 0.11 \\   
\smalltablenr{7} & \textsc{RAMP$^-$} &  40 & \textbf{15.85} \small $\pm 0.02$ & $+$ 0.26 \\ 
\smalltablenr{8} & \textsc{RAMP$_{\delta_2}$}  & 40 &  \textbf{15.86} \small $\pm 0.04$ & $+$ 0.27 \\ 
\smalltablenr{9} & \textsc{RAMP$^-$-T}&40 &  \textbf{16.03}$^*$\small $\pm 0.02$ & $+$ 0.44 \\ 
\smalltablenr{10} &  \textsc{RAMP$_{\delta_2}$-T}& 40 & \textbf{15.84} \small $\pm 0.02$ & $+$ 0.25 \\ 
\bottomrule
\end{tabular}
\end{center}
\caption{BLEU scores for weakly supervised MT experiments. Boldfaced results are significantly better than the baseline at $p<0.05$ according to \texttt{multeval} \citep{clark2011better}. 
$^*$ marks a significant difference over \textsc{RAMP$^-$}. 
}
\label{exp:weakmt_results}
\end{table}

\paragraph{Experimental Results.}
Results for the different objectives can be found in Table \ref{exp:weakmt_results}. 
The ramp losses \textsc{RAMP}, \textsc{RAMP1$^-$} and \textsc{RAMP2}, which do not incorporate bipolar supervision from $d^+$ \textit{and} $d^-$ (lines 2, 3 and 4) actually deteriorate in performance. This shows that supervision from only $d^+$ or only $d^-$ is insufficient. The deteriorating effect is strongest for \textsc{RAMP}, which  uses $d^+$ to select both $y^+$ and $y^-$. 
We explain this by the fact that $d^+$ is an imperfect label. Trying to push the model to perfectly reproduce $d^+$ will not lead to a good translation. 
The same observation holds true for MRT$_{\delta_1}$. This objective only includes the reward  $\delta_1(y,d^+)$. Compared to the RAMP objectives, the decrease for MRT$_{\delta_1}$ is smaller. 

On the other hand, MRT$_{\delta_2}$, which  incorporates bipolar supervision, produces a nominal improvement over the MLE baseline.
This objective is outperformed by \textsc{RAMP$^-$} and  \textsc{RAMP$_{\delta_2}$}. Both objectives produce a small, but significant, improvement of 0.3\% BLEU over the MLE baseline. This result shows that bipolar supervision is crucial for success in this weak supervision scenario. It also shows that unlike MRT, for the bipolar ramp loss it does not matter whether $\delta_1$ or $\delta_2$ is used, as they both capture the same idea. The superiority of these objectives over MRT shows again the success of intelligently selecting positive and negative outputs.  
Another small, but significant improvement is produced by  the token-level variant \textsc{RAMP$^-$-T}, leading to the best overall result. 

To summarize, we find that for this task, which uses very weak supervision from document-level links, small improvements can be obtained. To achieve these improvements, it is imperative to employ objectives which include bipolar supervision from $d^+$ and $d^-$.  This finding holds for both ramp loss and MRT. The best overall result is obtained using ramp loss in the token-level variant.

\paragraph{Analysis of Translation Results.}

\begin{figure}[t]
\includegraphics[scale=0.5]{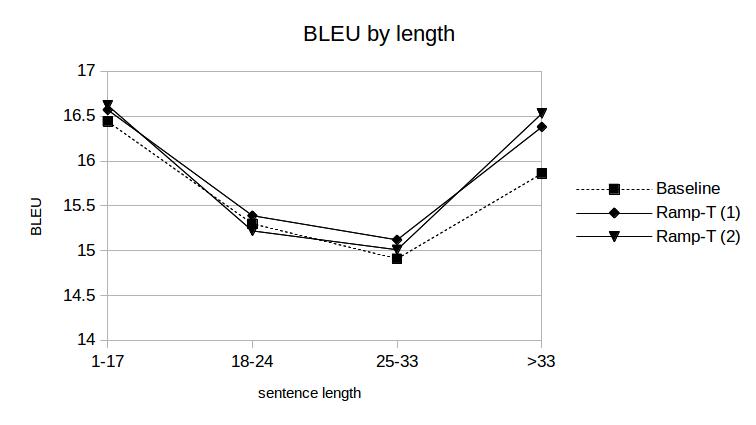}
\caption{BLEU scores by sentence length for the MLE Baseline and the \textsc{RAMP$^-$-T} runs.}
\label{fig:bleu_by_len}
\end{figure}

As the improvements in the translation experiments are very small, we conduct a small-scale analysis to better determine the nature of the gains. Our analysis is inspired by \newcite{bentivogli2016neural}. We compare the weakly supervised MLE baseline to the best experiment in this setting, which uses the bipolar token-level ramp loss \textsc{RAMP$^-$-T}. 

We first analyze the performance by sentence length. We separate the translations into source length brackets and score each bracket separately. The brackets represent quartiles of the source length distribution, ensuring an approximately equal amount of sentences in each bracket. Results are shown in Figure \ref{fig:bleu_by_len}. For all systems, we observe a drop in performance up to an input length of 33. Surprisingly, BLEU scores increase again for the top bracket (source length $>33$). For this bracket, we also see the biggest gap between MLE and  \textsc{RAMP$^-$-T} of 0.52 and 0.67\% BLEU for the two runs. This increase is mitigated by much weaker increases in the bottom brackets. A possible explanation for the weaker performance of MLE in the top bracket is the observation that hypotheses produced by the MLE system are longer than for \textsc{RAMP$^-$-T}. For the top bracket, hypothesis lengths exceed reference lengths for all systems. However, for MLE this over-generation is more severe at 106\% of the reference length compared to \textsc{RAMP$^-$-T} at 102\%, potentially causing a higher loss in precision. 

\begin{figure}[t]
\includegraphics[scale=0.5]{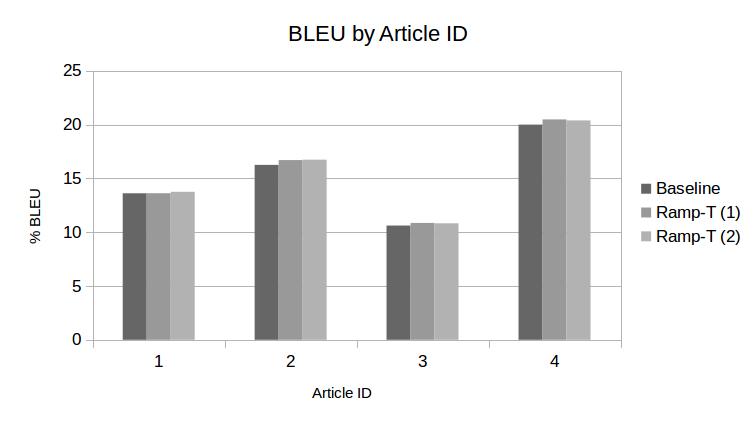}
\caption{BLEU scores by Wikipedia article for the MLE Baseline and the \textsc{RAMP$^-$-T} runs.}
\label{fig:bleu_by_article}
\includegraphics[scale=0.5]{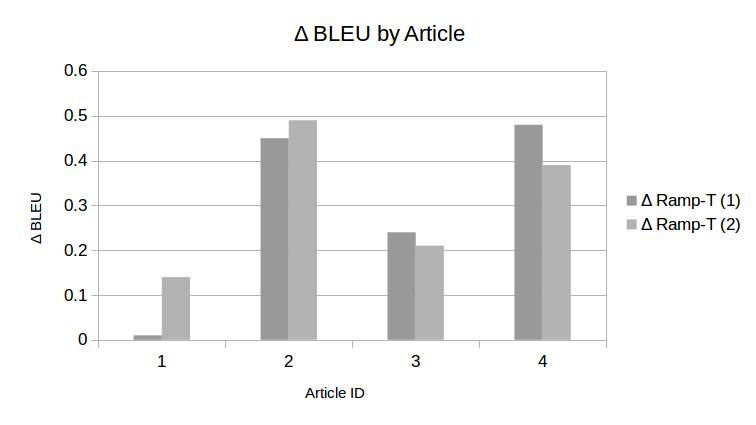}
\caption{Improvements in BLEU scores by Wikipedia article for the \textsc{RAMP$^-$-T} runs.}
\label{fig:delta_by_article}
\end{figure}

As our test set consists of parallel sentences extracted from four Wikipedia articles, we can examine the performance for each article separately. 
Figure \ref{fig:bleu_by_article} shows the results. We observe large differences in performance according to article ID. These are probably caused by some articles being more similar to the out-of-domain training data than others.  Comparing \textsc{RAMP$^-$-T} and MLE, we see that \textsc{RAMP$^-$-T} outperforms MLE for each article by a small margin.  Figure \ref{fig:delta_by_article} shows the size of the improvements by article. We observe that margins are bigger on articles with better baseline performance. This suggests that there are challenges arising from domain mismatch which are not addressed by our method. 

\begin{table*}[t]
\footnotesize
\renewcommand{\arraystretch}{1.2}
\begin{tabular}{lp{13.7cm}}
Source & Towards the end of the 19th century , a strong textile industry was developing itself in Sch\"uttorf with several large local businesses ( Schlikker \& S\"ohne , Gathmann \& Gerdemann , G. Sch\"umer \& Co. and ten Wolde , later Carl Remy ; today 's RoFa is not one of the original textile companies , but was founded by H. Lammering and later taken over by Gerhard Schlikker jun. , Levert Rost and Wilhelm Edel ;\\ 
\textsc{MLE} &  Ende des 19. Jahrhunderts , eine starke Textilindustrie , \underline{die sich} in \underline{\textit{Ettorf}} mit mehreren gro{\ss}en lokalen Unternehmen ( Schlikker \& S\"ohne , Gathmann \& \underline{\textit{Ger\'eann}} , G. \underline{\textit{Schal}} \& Co. und \underline{\textit{zehn Wolde}} , sp\"ater Carl Remy ) \underline{entwickelt hat} ; die heutige RoFa ist nicht \underline{einer der urspr\"unglichen Textilunternehmen} , sondern wurde von H. Lammering \underline{\textit{[gegr\"undet]}} und sp\"ater von Gerhard \underline{\textit{Schaloker Junge}} , Levert Rost und Wilhelm Edel \"ubernommen .\\
\textsc{RAMP$^-$-T} & Ende des 19. Jahrhunderts \textbf{entwickelte sich} \underline{\textit{[in Sch\"uttorf]}} eine starke Textilindustrie mit mehreren gro{\ss}en lokalen Unternehmen ( Schlikker \& S\"ohne , Gathmann \& \textbf{Gerdemann} , G. \underline{\textit{Schal}} \& Co. und \underline{\textit{zehn Wolde}} , sp\"ater Carl Remy ; die heutige RoFa ist nicht \textbf{eines der urspr\"unglichen Textilunternehmen} , sondern wurde von H. Lammering \underline{\textit{[gegr\"undet]}} und sp\"ater von Gerhard \underline{\textit{Schaloker Junge}} , Levert Rost und Wilhelm Edel \"ubernommen .\\
Reference & gegen Ende des 19. Jahrhunderts entwickelte sich in Sch\"uttorf eine starke Textilindustrie mit mehreren gro{\ss}en lokalen Unternehmen ( Schlikker \& S\"ohne , Gathmann \& Gerdemann , G. Sch\"umer \& Co. und ten Wolde , sp\"ater Carl Remy , die heutige RoFa ist keine urspr\"ungliche Textilfirma , sondern wurde von H. Lammering gegr\"undet und sp\"ater von Gerhard Schlikker jun. , Levert Rost und Wilhelm Edel \"ubernommen .)\\
\end{tabular}

\caption{MT example from Article 2 in the test set. All translation errors are \underline{underlined}. Incorrect proper names are also set in \underline{\textit{cursive}}. Omissions are inserted in brackets and set in cursive \underline{\textit{[like this]}}. Improvements by \textsc{RAMP$^-$-T} over MLE are marked in \textbf{boldface}.}
\label{tab:mt_example}
\end{table*}

Lastly, we present an examination of example outputs. Table \ref{tab:mt_example} shows an example of a long sentence from Article 2, which describes the German town of Sch\"uttorf. This article is originally in German, meaning that our model is back-translating from English into German. The reference contains some awkward or even ungrammatical phrases such as \textit{``was developing itself''}, a literal translation from German. 
The example also illustrates that translating Wikipedia involves handling frequent proper names (there are 11 proper names in the example). 
Both models struggle with translating proper names, but \textsc{RAMP$^-$-T} produces the correct phrase \textit{``Gathmann \& Gerdemann''}, while MLE fails to do so. The  \textsc{RAMP$^-$-T} translation is also fully grammatical, while MLE incorrectly translates the main verb phrase \textit{``was developing itself''} into a relative clause, and contains an agreement error in the translation of the noun phrase \textit{``one of the original textile companies''}.
While making fewer errors in grammar and proper name translation, \textsc{RAMP$^-$-T} contains two deletion errors while MLE only contains one. This could be caused by the active optimization of sentence length in the ramp loss model. 

\section{Fully Supervised Machine Translation} 
While our work focuses on weakly supervised tasks, we also conduct experiments using a fully supervised MT task. These experiments are motivated on the one hand by adapting the findings of \citet{gimpel2012ramp} to the neural MT paradigm, and on the other hand to expand the work by \citet{edunov2018classical} on applying classical structured prediction losses to neural MT. 

\begin{table*}
\begin{center}
\begin{tabular}{llp{6.5cm}}
\toprule
 Loss & $y^+$ & $y^-$  \\
 \midrule
RAMP & $\argmax_{y} \pi_w(y|x) - \alpha( 1 - \operatorname{BLEU}_{+1}(y,\bar{y}))$  & $\argmax_{y} \pi_w(y|x) + \alpha( 1 - \operatorname{BLEU}_{+1}(y,\bar{y}))$  \\ 
RAMP1 & $\hat{y}$  & $\argmax_{y} \pi_w(y|x) + \alpha ( 1 - \operatorname{BLEU}_{+1}(y,\bar{y}))$   \\
RAMP2 & $\argmax_{y} \pi_w(y|x) - \alpha( 1 - \operatorname{BLEU}_{+1}(y,\bar{y}))$   & $\hat{y}$  \\ 
PERC1 & $ \bar{y} $   & $\hat{y}$   \\ 
PERC2 & $\argmax_{y}\operatorname{BLEU}_{+1}(y,\bar{y}) $  & $\hat{y}$  \\ 
\bottomrule 
\end{tabular}
\caption{Configurations for $y^+$ and $y^-$ for fully supervised MT. $\hat{y}$ is the highest-probability model output, $\bar{y}$ is a gold standard reference.  $\pi_w(y|x)$ is the probability of $y$ according to the model. The $\argmax_y$ is  taken over the $k$-best list $\mathcal{K}(x)$. $\operatorname{BLEU}_{+1}$  is smoothed per-sentence BLEU and $\alpha$ is a scaling factor.} 
\label{tab:ramp_fullmt}
\end{center}
\end{table*}

\paragraph{Ramp Loss Objectives.}
For fully supervised MT we assume access to one or more reference translations $\bar{y}$ for each input $x$. 
The reward $\operatorname{BLEU}_{+1}(y,\bar{y})$ is a per-sentence approximation of the BLEU score.\footnote{We use the BLEU score with add-1 smoothing for $n>1$ as proposed by \citet{chen2014smoothing}. }
Table \ref{tab:ramp_fullmt}
shows the different definitions of $y^{+}$ and $y^{-}$, which give rise to different ramp losses. 
RAMP, RAMP1, and RAMP2 are defined analogously to the other tasks. We again include a hyperparameter $\alpha > 0$ interpolating cost function and model score when searching for $y^+$ and $y^-$. 
\citet{gimpel2012ramp} also include the perceptron loss in their analysis. 
PERC1 is a re-formulation of the Collins perceptron \citep{collins2002discriminative} where the reference is used as $y^+$ and $\hat{y}$ is used as $y^-$. A comparison with PERC1 is not possible for the weakly supervised tasks in the previous sections, as gold structures are not available for these tasks. 
With neural MT and subword methods we are able to compute this loss for any reference without running into the problem of \textit{reachability} that was faced by phrase-based MT \citep{LiangETAL:06}.  However, using sequence-level training towards a reference can lead to degenerate solutions where the model gives low probability to all its predictions \citep{shen2016minimum}. PERC2 addresses this problem by replacing $\bar{y}$ by a surrogate translation which achieves the highest \textsc{BLEU}$_{+1}$ score in $\mathcal{K}(x)$. This approach is also used by \citet{edunov2018classical} for the loss functions which require an oracle. PERC1 corresponds to equation (9), PERC2 to equation (10) of \citep{gimpel2012ramp}.

\paragraph{Experimental Setup.}
We conduct experiments on the IWSLT 2014 German-English task, which is based on \citet{cettolo2012wit3} in the same way as \citet{edunov2018classical}. The training set contains 160K sentence pairs. We set the maximum sentence length to 50 and use BPE with 14,000 merge operations. \citet{edunov2018classical} sample 7K sentences from the training set as heldout data. We do the same, but only use 1/10th of the data as heldout set to be able to validate often. 

Our baseline system (MLE) is a BiLSTM encoder-decoder with attention, which is trained using the MLE objective. Word embedding and hidden layer dimensions are set to 256. 
We use batches of 64 sentences for baseline training and batches of 40 inputs for training \textsc{RAMP} and \textsc{PERC} variants. MRT makes an update after each input using all sampled outputs and resulting in a batch size of 1.  
All experiments use dropout for regularization, with dropout probability set to 0.2 for embedding and hidden layers and to 0.1 for source and target layers. 
During MLE-training, the model is validated every 2500 updates and training is stopped if the MLE loss on the heldout set worsens for 10 consecutive validations.

For metric-augmented training, we use SGD for optimization with learning rates optimized on the development set.
Ramp losses and PERC2 use a $k$-best list of size 16. For ramp loss training, we set $\alpha=10$. RAMP and PERC variants both use a learning rate of 0.001.  A new $k$-best list is generated for each input using the current model parameters.
We compare ramp loss to MRT as described above. For MRT, we use SGD with a learning rate of 0.01 and set $S=16$ and $S^\prime=10$. As \citet{edunov2018classical} observe beam search to work better than sampling for MRT, we also run an experiment in this configuration, but find no difference between results. As beam search runs significantly slower, we only report sampling experiments.

The model is validated on the development set after every 200 updates for experiments with batch size 40 and after 8,000 updates for MRT experiments with batch size 1. The stopping point is determined by the BLEU score on the heldout set after 25 validations.  As we are training on the same data as the MLE baseline, we also apply dropout during ramp loss training to prevent overfitting. 
BLEU scores are computed with Moses' \texttt{multi-bleu.perl} on tokenized, truecased output. Each experiment is run 3 times and results are averaged over the runs. 

\paragraph{Experimental Results.}
As shown in Table \ref{exp:fullmt_results}, all experiments except for PERC1 yield improvements over MLE, confirming that sequence-level losses which update towards the reference can lead to degenerate solutions. 
For MRT, our findings show similar performance to the initial experiments reported by \citet{edunov2018classical} who gain 0.24 BLEU points on the same test set.\footnote{See their Table 2. Using interpolation with the MLE objective, \citet{edunov2018classical} achieve $+$0.7 BLEU points. As we are only interested in the effect of sequence-level objectives, we do not add MLE interpolation. The best model by \citet{edunov2018classical} achieved a BLEU score of 32.91\%. It is possible that these score are not directly comparable to ours due to different pre- and post-processing. They also use a multi-layer CNN architecture \citep{gehring2017convolutional}, which has been shown to outperform a simple RNN architecture such as ours.} 
PERC2 and RAMP2 improve over the MLE baseline and PERC1,
but perform on a par with MRT and each other. 
Both RAMP and RAMP1 are able to outperform MRT, PERC2 and RAMP2, 
with the bipolar objective RAMP also outperforming RAMP1 by a narrow margin. 
The main difference between RAMP and RAMP1, compared to PERC2 and RAMP2, is the fact that the latter objectives use $\hat{y}$ as $y^-$, while the former use a \textit{fear} translation with high probability and low BLEU$_{+1}$. 
We surmise that for this fully supervised task, selecting a $y^-$ which has some known negative characteristics is more important for success than finding a good $y^+$. \textsc{RAMP}, which fulfills both criteria, still outperforms \textsc{RAMP2}. This result re-confirms the superiority of bipolar objectives compared to non-bipolar ones.
While still improving over MLE, token-level ramp loss \textsc{RAMP-T} is outperformed by \textsc{RAMP} by a small margin. 
This result suggests that when employing a metric-augmented objective on top of an MLE-trained model in a full supervision scenario without domain shift, there is little room for improvement from  token-level supervision, while gains can still be obtained from additional sequence-level information  captured by the external metric, such as information about the sequence length.

To summarize, our findings on a fully supervised task show the same small margin for improvement as \citet{edunov2018classical}, without any further tuning of performance, e.g.~by interpolation with the MLE objective. Bipolar RAMP is found to outperform the other losses. This observation is also consistent with the results by \citet{gimpel2012ramp} for phrase-based MT. We conclude that for fully supervised MT, deliberately selecting a \textit{hope} and \textit{fear} translation is beneficial.

\begin{table}[t]
\begin{center}
\begin{tabular}{cllll}
\toprule
& & $M$ & \% BLEU & $\Delta$ \\
\midrule
\smalltablenr{1} & MLE & 64 & 31.99 \\ 
\smalltablenr{2} & \textsc{MRT} & 1 & \textbf{32.17} \small $\pm$ 0.02 & $+$ 0.18 \\ 
\smalltablenr{3} & \textsc{PERC1} & 40 &  31.91 \small $\pm$ 0.02 & $-$ 0.08 \\ 
\smalltablenr{4} & \textsc{PERC2} &  40 & \textbf{32.22} \small $\pm$ 0.03 & $+$ 0.23 \\ 
\smalltablenr{5} & \textsc{RAMP1} & 40 & \textbf{32.36}$^*$ \small $\pm$ 0.05 & $+$ 0.37\\ 
\smalltablenr{6} & \textsc{RAMP2}  &  40 & \textbf{32.19} \small $\pm$ 0.01 & $+$ 0.20 \\ 
\smalltablenr{7} & \textsc{RAMP} & 40 & \textbf{32.44}$^{**}$ \small $\pm$ 0.00 & $+$ 0.45\\ 
\smalltablenr{8} & \textsc{RAMP-T} & 40 & \textbf{32.33}$^{*}$ \small $\pm$ 0.00 & $+$ 0.34\\ 
\bottomrule
\end{tabular}
\end{center}
\caption{BLEU scores for fully supervised MT experiments. Boldfaced results are significantly better than MLE at $p < 0.01$ according to \texttt{multeval} \citep{clark2011better}. $^*$ marks a significant difference to \textsc{MRT} and \textsc{PERC2}, and $^{**}$ marks a difference to \textsc{RAMP1}.
}
\label{exp:fullmt_results}
\end{table}

\section{Conclusion} 
We presented a study of weakly supervised learning objectives for three neural sequence-to-sequence learning tasks. 
In our first task of semantic parsing, question-answer pairs provide a weak supervision signal to find parses that execute to the correct answer. We show that ramp loss can outperform MRT if it incorporates bipolar supervision where parses that receive negative feedback are actively discouraged.
The best overall objective is constituted by the token-level ramp loss.
Next, we turn to weak supervision for machine translation in form of cross-lingual document-level links. We present two ramp loss objectives which combine bipolar weak supervision from a linked document $d^+$ and an irrelevant document $d^-$. 
Again, the bipolar ramp loss objectives outperform MRT, and the best overall result is obtained using token-level ramp loss. 
Finally, to tie our work to previous work on supervised machine translation, we conduct experiments in a fully supervised scenario where gold references are available and a metric-augmented loss is desired to reduce the exposure bias and the loss-evaluation mismatch. 
Again, the bipolar ramp loss objective performs best, but we find that the overall margin for improvement is small without any additional engineering. 
We conclude that ramp loss objectives show promise for neural sequence-to-sequence learning, especially when it comes to weakly supervised tasks where the MLE objective cannot be applied. 
In contrast to ramp losses that either operate only in the undesirable region of the search space (``cost-augmented decoding'' as in \textsc{RAMP1}) or only in the desirable region of the search space (``cost-diminished decoding'' as in \textsc{RAMP2}), bipolar \textsc{RAMP} operates in both regions of the search space when extracting supervision signals from weak feedback. We showed that MRT can be turned into a bipolar objective by defining a metric that assigns negative values to bad outputs. This improves the performance of MRT objectives. However, the ramp loss objective is still superior as it is easy to implement and efficient to compute. Furthermore, on weakly supervised tasks our novel token-level ramp loss objective \textsc{RAMP-T} can obtain further improvements over its sequence-level counterpart because it can more directly assess which tokens in a sequence are crucial to its success or failure.

\section*{Acknowledgments}
The research reported in this paper was supported in part by DFG grant RI-2221/4-1. We would like to thank the reviewers for their helpful comments. 

\newpage
\bibliography{lit}
\bibliographystyle{acl_natbib}

\end{document}